%% file: main.tex
\documentclass[10pt,twocolumn,letterpaper]{article}

\usepackage{cvpr}      % To produce the REVIEW version
\usepackage{graphicx} % Required for inserting images
\usepackage{multirow}
\usepackage{placeins}
\usepackage{amsmath,amssymb,amsfonts}
\usepackage{algorithm}
\usepackage{algorithmicx} % <===========================================
\usepackage{algpseudocode} 
\usepackage{array}
\usepackage{comment}
\usepackage{pifont}% http://ctan.org/pkg/pifont
\newcommand{\cmark}{\ding{51}}%
\newcommand{\xmark}{\ding{55}}%
\input{preamble}

\definecolor{cvprblue}{rgb}{0.21,0.49,0.74}
\usepackage[pagebackref,breaklinks,colorlinks,citecolor=cvprblue]{hyperref}

\newcommand{\Method}{\textsc{RankED}}

\usepackage{mathtools} % for xarrow
\usepackage{color}
\usepackage{arydshln} % cdashline

\newcommand{\pred}{{{p}}}
\newcommand{\predsingle}{{{p}}}
\newcommand{\target}{{{y}}}
\newcommand{\targetsingle}{{{y}}}
\newcommand{\Unc}{c} %{\mathcal{U}}

\definecolor{rankcolor}{RGB}{32, 51, 20}
\definecolor{sortcolor}{RGB}{26, 44, 76}

\def\paperID{2469} % *** Enter the Paper ID here
\def\confName{CVPR}
\def\confYear{2024}

\title{RankED: Addressing Imbalance and Uncertainty in Edge Detection\\ Using Ranking-based Losses}

\author{Bedrettin Cetinkaya, Sinan Kalkan$^\dag$, Emre Akbas$^\dag$\\
Department of Computer Engineering and METU ROMER Robotics Center\\
Middle East Technical University, Ankara, Turkey\\
{\tt\small \{bckaya, skalkan, eakbas\}@metu.edu.tr}
}
\newcommand{\CStart}[0]{\begin{center}  \begin{tabular}{@{}c@{}}}
\newcommand{\CEnd}[0]{\end{tabular} \end{center}}
\newcommand{\kaynak}[1]{\hfill\textcolor{gray}{#1}}

\begin{document}

\maketitle
\begin{abstract}
Detecting edges in images suffers from the problems of (P1) heavy imbalance between positive and negative classes as well as (P2) label uncertainty owing to disagreement between different annotators. Existing solutions address P1 using class-balanced cross-entropy loss and dice loss and P2 by only predicting edges agreed upon by most annotators. In this paper, we propose RankED, a unified ranking-based approach that addresses both the imbalance problem (P1) and the uncertainty problem (P2). RankED tackles these two problems with two components: One component which ranks positive pixels over negative pixels, and the second which promotes high confidence edge pixels to have more label certainty. We show that RankED
outperforms previous studies and sets a new state-of-the-art on NYUD-v2, BSDS500 and Multi-cue datasets. Code is available at \url{https://ranked-cvpr24.github.io}. 
\end{abstract}

\def\thefootnote{\dag}\footnotetext{Equal contribution.}

\input{sect_introduction}

\input{sect_relatedWork}
\input{sect_method}
\input{sect_experiments}
\input{sect_conclusion}

\section*{Acknowledgements} We gratefully acknowledge the computational resources  provided by METU-ROMER, Center for Robotics and Artificial Intelligence, Middle East Technical University. Dr. Akbas is supported by the ``Young Scientist Awards Program (BAGEP)" of Science Academy, Türkiye.

\FloatBarrier
\bibliographystyle{ieeenat_fullname}
\bibliography{refs}
%\FloatBarrier
%\input{sect_supp}

\end{document}

%% file: preamble.tex
\usepackage[dvipsnames]{xcolor}

%% file: sect_introduction.tex
\section{Introduction}
\begin{figure}
\centering
\includegraphics[width=0.81\linewidth]{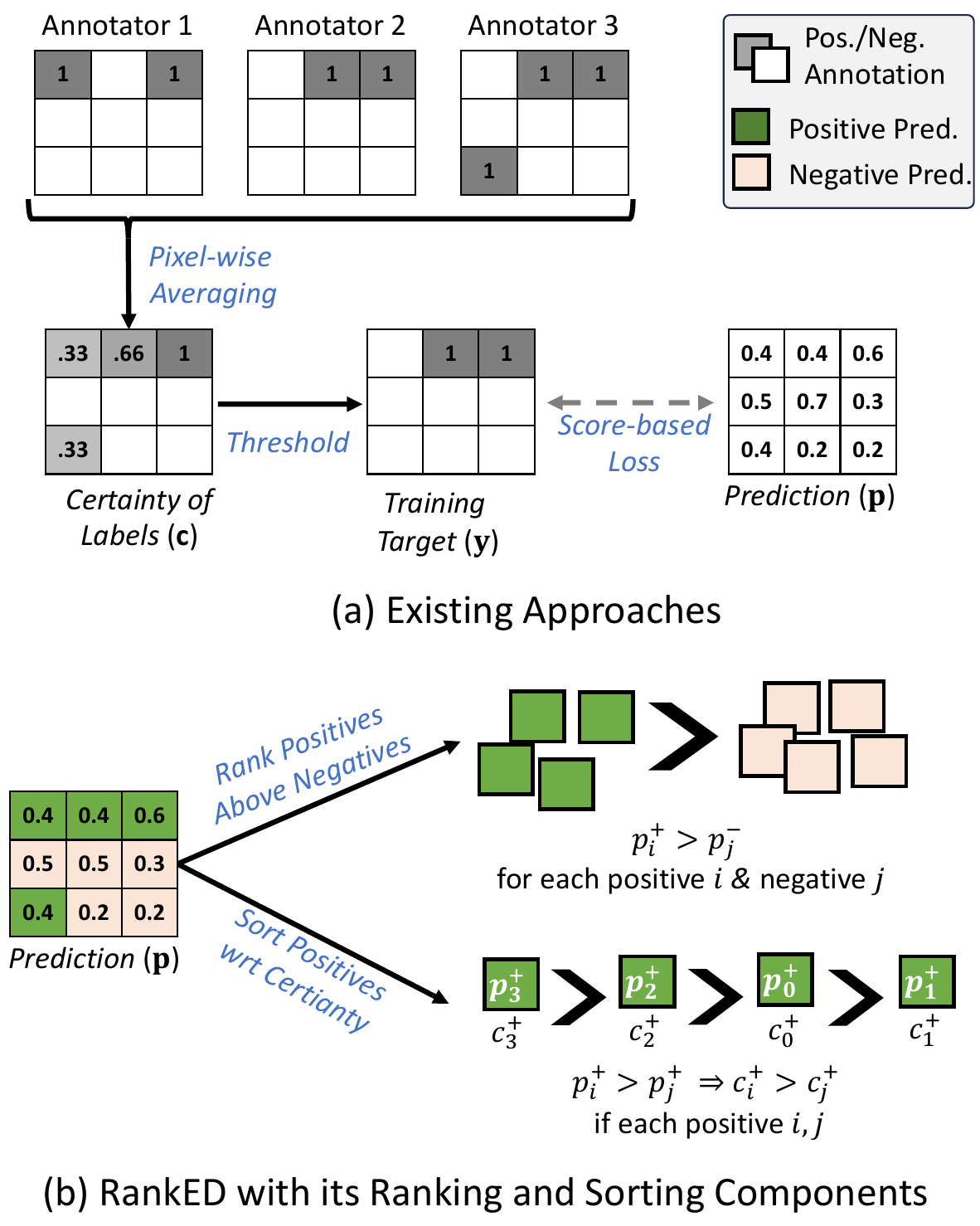}
  \caption{(a) Current approaches threshold label certainties and class-balanced cross-entropy loss for training edge detectors. (b) With RankED, we propose a unified approach which ranks positives over negatives to handle the imbalance problem and sorts positives with respect to their certainties.}
  %Overview of our method on a small example. Case (a) shows a multi-label of one RGB image. Case (b) shows label processing approaches for training. Case (c) shows a ranking task that aims to rank all edge pixels above each background pixel. Case (d) shows a sorting task that aims to sort edge pixels w.r.t proposed uncertainty-aware-label.}
  \label{fig:intro}
\end{figure}

Detecting contours of objects in a given image is a fundamental problem in Computer Vision. It has been approached as a machine learning problem since the introduction of the influential BSDS dataset \cite{arbelaez2010contour}. As with any learning-based approach, characteristics of the training data affects performance. One striking issue regarding ground-truth contour data is that contours are rare events. For example, in the BSDS dataset, only 7\% of all pixels within an image are marked as edge pixels\def\thefootnote{1}\footnote{Although some studies call such high-level, semantic edges as ``contour" and low-level edges as ``edge", we follow the recent literature \cite{su2021pixel,pu2022edter,deng2023learning,elharrouss2023refined,zhou2023treasure} and use the term ``edge" for contours in the rest of the paper.}. This creates a significant imbalance between the positive (edge) and negative (non-edge) classes, which hinders the training of machine learning models. Another important issue observed in edge data is the uncertainty regarding the ground-truth annotations. There exist a non-trivial amount of variation between the annotations produced by different human annotators, which essentially creates noise in the supervisory signal. 

These two issues of edge ground-truth data, namely the imbalance and uncertainty, have long been known, however, efforts to address them have remained limited. For the imbalance problem, although there is a vast literature on long-tailed and imbalance learning (see, e.g., \cite{zhang2023deep, wang2021review}), researchers have only explored using Dice Loss \cite{deng2018learning, deng2021learning, deng2023learning} and weighted cross-entropy loss \cite{xie2015holistically,liu2017richer, he2019bi,su2021pixel,pu2022edter}, which mitigate the problem to a certain extent. However, as we show in this paper, they are far from finding the optimal solution. The label uncertainty has been tackled using edge pixels agreed by most annotators \cite{xie2015holistically, liu2017richer, he2019bi,pu2022edter,deng2023learning} (Figure \ref{fig:intro}(a)). A recent exception is the study by Zhou et al. \cite{zhou2023treasure} who proposed jointly learning the mean and variances of labels using multi-variate Gaussian distributions. To the best of our knowledge, this is the only work in literature to attempt the uncertainty problem.

In this paper, we address both the imbalance and the uncertainty problems encountered in edge detection using ranking-based losses. We draw inspiration from the recent success of ranking-based losses in object detection \cite{APloss-PAMI20,rolinek2020optimizing,qian2020dr,oksuz2021rank} and instance segmentation \cite{oksuz2021rank}. Similar to edge detection, the data in these problems also manifest high imbalance between classes \cite{oksuz2020imbalance}. In particular, AP Loss \cite{APloss-PAMI20} and RS Loss \cite{oksuz2021rank} have shown to outperform focal loss and weighted cross-entropy in these problems. Their success stems from the ability to naturally balance the gradients for different classes \cite{aLRPLoss}. Inspired by these works, we propose \Method, which basically uses a ranking-based loss function to learn from data (Figure \ref{fig:intro}(b)). When a training image has only one annotation, then \Method~tries to rank edge pixels above non-edge pixels. When there are multiple annotations per training image, \Method~not only tries to rank edge pixels above non-edge pixels but also to sort edge pixels with respect to their annotation certainty. 

%Although the term ``contour" refers to high-level, semantic edges and the term ``edge" to low-level edges, the community seems to have preferred ``edge detection" to refer to the problem of learning-based object contour detection. We follow this trend and use ``edge detection". 
 
\noindent\textbf{Contributions.} Our main contributions are as follows: 

\begin{itemize}
\item We propose \Method, a new ranking-based loss function for edge detection.
\item \Method~simultaneously addresses the imbalance and uncertainty issues commonly encountered in edge detection datasets. 
\item Our experiments on three edge detection datasets (BSDS, NYUDv2, MultiCue) show that when integrated with Swin-Transformer, \Method~consistently outperforms all SOTA models in average precision (AP).  
\end{itemize}

%% file: sect_relatedWork.tex
\section{Related Work}

%We first provide a review of relevant studies.

\noindent \textbf{Edge Detection.}
Before the rise of deep learning, the conventional edge detection methods were based on hand-crafted filters such as Sobel \cite{kittler1983accuracy}, Canny \cite{canny1986computational}, and Laplacian of Gaussian \cite{marr1980theory}. Deep learning has enabled detecting edges directly from data and has provided high quality results \cite{xie2015holistically,bertasius2015high,deng2018learning,kelm2019object,he2019bi,kokkinos2015pushing,maninis2016convolutional,pu2021rindnet,shen2015deepcontour,su2021pixel,xu2017learning,pu2022edter,zhou2023treasure}.

Earlier deep learning based methods \cite{bertasius2015deepedge, shen2015deepcontour, bertasius2015high} used features extracted from fixed patches. In this approach, CNNs detect edges using extracted patches around candidate contour points. Following studies extended this patch-based approach with end-to-end learning. These studies focused on detecting edges at multiple scales \cite{xie2015holistically} by combining features at different layers \cite{liu2017richer} and in cascades \cite{he2019bi,elharrouss2023refined} and making these networks faster \cite{su2021pixel}. Recently, attention-based modules and transformers have been shown to provide significant improvements \cite{pu2022edter,deng2023learning}. %\SK{Asagida comment out ettigim yerden calismalari yukariya dagitabilir misin? \BC{referansları ekledim hocam dediğiniz gibi.}}

%For example, HED \cite{xie2015holistically} introduced an architecture that learns multi-scale and multi-level features together. In this model, labels are used as supervision signals for the outputs of all blocks. RCF \cite{liu2017richer} improves HED's architecture and uses richer features by combining the output of all convolution layers. BDCN \cite{he2019bi}  exploits the advantages of a bi-directional cascade structure via layer-specific supervision signals. PiDiNet \cite{su2021pixel} exhibits lightweight architecture that uses pixel difference convolutions. RINDNET \cite{pu2021rindnet} processes different types of edges with separate branches. EDTER \cite{pu2022edter} introduces the first transformer-based model for edge detection. Also, CHRNet \cite{elharrouss2023refined} refines edge to remove erroneous edges via cascaded CNN. ACTD \cite{deng2023learning} introduced a contour Transformer network that learns sharp and crisp object boundaries.

\noindent \textbf{Loss Functions for Edge Detection.}
\label{sect:rel_work_loss_functions} Edge detection is generally tackled using score-based classification losses, e.g., Cross Entropy (CE) Loss \cite{xie2015holistically,he2019bi,su2021pixel,pu2022edter}. However, owing to the severe imbalance problem between positives ($\mathcal{P}$) and negatives ($\mathcal{N}$), CE Loss is often weighted with a class balancing (CB) factor, e.g. as follows \cite{xie2015holistically}:
\begin{equation}
\label{eq:wbce}
\mathcal{L}_{CB}=\sum_i -\beta  \target_i \log(\pred_i) -(1-\beta)(1 -\target_i) \log(1 -\pred_i),
\end{equation} 
where $\pred_i$ is the edge prediction probability for pixel $i$ and $\target_i$ is its target. The terms are multiplied with weights to mitigate the imbalance issue: $\beta$ = $ |\mathcal{N}| / |\mathcal{N}\cup \mathcal{P}|$, $(1-\beta)$ = $ |\mathcal{P}| / |\mathcal{N}\cup \mathcal{P}|$. With these weights, class-balanced loss function applies a higher penalty for edge pixels as they are rare. 

A more common approach is to use an adaptation of the Dice Loss \cite{sudre2017generalised} for edge detection \cite{deng2018learning,deng2021learning,deng2023learning}:
\begin{equation}
\label{eq:dice}
\mathcal{L}_{DICE}=\frac{\sum_i \predsingle_i^2+\sum_i \targetsingle_i^2}{2 \sum_i \predsingle_i \targetsingle_i},
\end{equation} 
where $i$ runs over all pixels. Dice Loss is often combined with CE Loss \cite{deng2018learning, deng2021learning, deng2023learning}:
\begin{equation}
\label{eq:dice_CE}
\mathcal{L}_{final} = \alpha \mathcal{L}_{DICE} + \beta \mathcal{L}_{CE} ,
\end{equation} 
where $\alpha$ and $\beta$ balance the two loss functions.

Although there are studies that propose regularization terms to improve edge detection quality (e.g., sharpness \cite{deng2023learning}); Cross Entropy, Class Balanced Cross Entropy and Dice are the de facto losses used for training deep edge detectors. To the best of our knowledge, there are no ranking-based losses proposed as an alternative.

\noindent \textbf{Using Uncertainty in Edge Detection.} Although uncertainty has been shown to be a useful measure for various Computer Vision problems such as classification \cite{kendall2017uncertainties}, object detection \cite{kraus2019uncertainty, harakeh2020bayesod}, depth estimation \cite{poggi2020uncertainty,hornauer2022gradient},  and semantic segmentation \cite{huang2018efficient, zheng2021rectifying}, there is only one study \cite{zhou2023treasure} employing uncertainty in edge detection. Zhou et al. \cite{zhou2023treasure} propose an uncertainty-aware edge detector (UAED) that exploits the multi-label nature of the edge detection problem. Their method jointly learns the mean and variance of given inputs and constructs multi-variate Gaussian distribution using predicted mean and variance. Moreover, their method gives more importance to the loss of pixels with higher uncertainty.  %use the uncertainty of edges to determine the importance of pixels, i.e. more priority for higher uncertainty. \SK{Bedrettin, uncertainty ile neyi yonlendirdiklerini, neyi promote ettiklerini de ekleyebilir misin? \BC{Ekledim hocam.} }

\begin{figure*}[hbt!]
\centering\footnotesize
\includegraphics[width=0.76\linewidth]{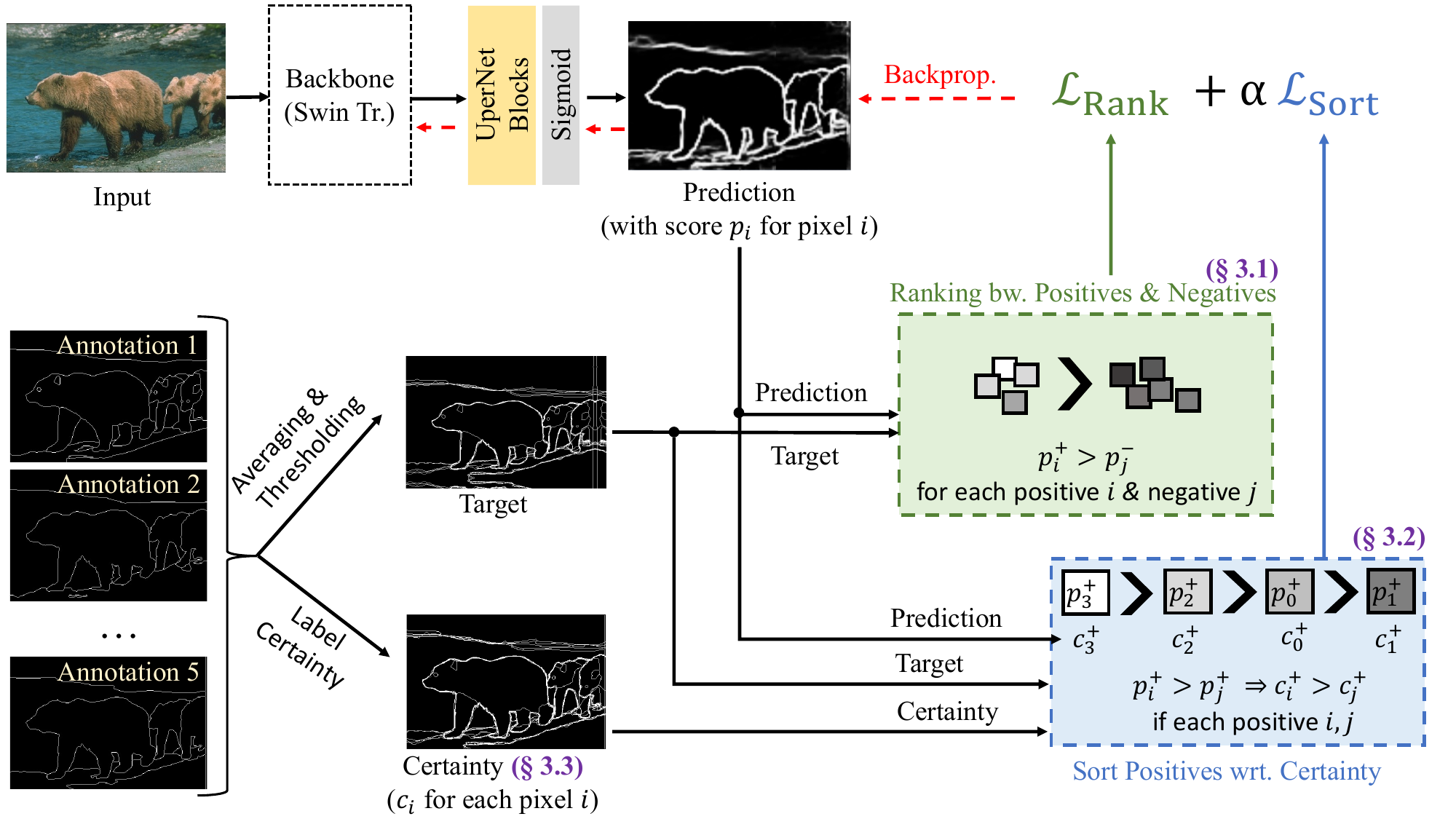}
  \caption{An overview of \Method. \Method~introduces two novel loss functions for edge detection: \textcolor{rankcolor}{$\mathcal{L}_{\textrm{Rank}}$} ($\S$ \ref{sect:ranking}) for ranking positive pixels over background pixels, and \textcolor{sortcolor}{$\mathcal{L}_{\textrm{Sort}}$} ($\S$ \ref{sect:sorting}) for sorting positive pixels with respect to their (un)certainties ($\S$ \ref{sect:uncertainty}).}
  \label{fig:model}
\end{figure*}

\noindent \textbf{Comparative Summary} We note from the literature reviewed above and their summary in Table \ref{tab:comparative_summary} that there are no studies that adopt a ranking approach to edge detection. There is only one study \cite{zhou2023treasure} utilizing label uncertainty in edge detection, which, however, does not use a ranking-based approach. %\SK{Bedrettin, cumleyi tamamlayabilir misin?}.

\begin{table}[hbt!]
    \centering
    
    \resizebox{0.9\columnwidth}{!}{%
    \begin{tabular}{|c|cc|}\hline
        \multirow{2}{*}{Method} & Ranking-based & \multirow{2}{*}{Uncertainty?}\\ 
          &  Loss Function?  &  \\ \hline\hline
        BDCN \cite{he2019bi} \kaynak{(CVPR '19)} &  \xmark &   \xmark\\        
        PiDiNet \cite{su2021pixel} \kaynak{(ICCV '21)} &  \xmark &   \xmark\\ 
        EDTER \cite{pu2022edter} \kaynak{(CVPR '22)} &  \xmark &   \xmark\\ 
        FCL-Net \cite{xuan2022fcl} \kaynak{(NN '22)} &  \xmark &   \xmark\\ 
        ACTD \cite{deng2023learning} \kaynak{(NeuroCom. '23)} & \xmark & \xmark \\ 
        CHRNet \cite{elharrouss2023refined} \kaynak{(Pat.Rec. '23)} & \xmark & \xmark \\ 
        UAED \cite{zhou2023treasure} \hfill \kaynak{(CVPR '23)}  &  \xmark & 
        \cmark \\ \hline
        \Method~ \hfill \kaynak{(This paper)} & \cmark & \cmark \\ \hline
    \end{tabular}
    }
    \caption{A comparative summary with recent studies. \label{tab:comparative_summary}}
\end{table}

%% file: sect_method.tex
\section{Methodology: \Method}

%\subsection{Overview}

\Method~is a ranking-based solution with two novel components for edge detection (Figure \ref{fig:model}):

\begin{itemize}
    \item Ranking Positives over Negatives ($\mathcal{L}_{\textrm{Rank}}$): \Method~is supervised with a loss function that ranks the classification scores of positives (edge pixels) over negatives (non-edge pixels). For this, we use a differentiable approximation of Average Precision as a loss function, following its successful applications in object detection \cite{chen2019towards}.

    \item Sorting Positives with Respect to their Uncertainties ($\mathcal{L}_{\textrm{Sort}}$): Positive pixels tend to have different uncertainties, which can be used to adjust their supervision. \Method~gives more priority for positive pixels with higher certainty. For this, we adapt the loss function proposed by Oksuz et al. \cite{oksuz2021rank} for sorting positives with respect to their localization qualities in object detection.
\end{itemize}

%We use a base model of Swin Transformer \cite{liu2021swin} and UPerNet \cite{xiao2018unified} as a backbone and prediction head, respectively.  Our purpose is to explore ranking-based loss functions in the context of edge detection, so we designed the model as simple as possible and did not use any side outputs. The overview of our model is shown in the upper part of Figure \ref{fig:model}.

%Then, we integrate ranked-based loss functions for two scenarios: (i) Single-label (AP-Loss) and (ii) Multi-label (RS-Loss) cases, as shown in the lower part of Figure \ref{fig:model}.

%\begin{figure}[hbt!]
%\centering
%\includegraphics[width=1\linewidth]%{figures/imbalance.pdf}
%  \caption{The heavy imbalance between background %and edge pixels  within the BSDS and Multicue datasets.}
%  \label{fig:imbalance}
%\end{figure}

\subsection{Ranking Positive Edge Pixels over Negatives} %Single-label Case: AP-Loss}
\label{sect:ranking}

To rank the scores of positive edge pixels higher than those of the negative pixels, we adapt the ranking-based loss function proposed by Chen et al. \cite{chen2019towards} for object detection. Chen et al. approximated Average Precision as a differentiable ranking based loss function, which was later shown by Oksuz et al. \cite{aLRPLoss} to be very robust to extreme positive-negative imbalance. As edge detection exhibits a severe imbalance between positive and negative classes (positive pixels are 1\%, 3\%, 7\% of all pixels for Multicue-boundary, Multi-cue edge, and BSDS datasets, respectively), a ranking-based loss maximizing Average Precision (AP) can be beneficial for edge detection. 

We define $\mathcal{L}_{\textrm{Rank}}$, the loss for ranking positive edge pixels above negatives in terms of Average Precision by counting positives and negatives as follows (following the notation introduced in Section \ref{sect:rel_work_loss_functions} and Chen et al. \cite{APloss-PAMI20}):
\begin{equation}\label{eqn:ap_loss}
\mathcal{L}_{\textrm{Rank}}=1-\mathrm{AP}=1-\frac{1}{|\mathcal{P}|} \sum_{i \in \mathcal{P}} \frac{\operatorname{rank}^{+}(i)}{\operatorname{rank}(i)}, \\
\end{equation}
{\noindent}where $\operatorname{rank}^{+}(i)$ is the rank of pixel $i$ among positives ($\mathcal{P}$) defined as $\operatorname{rank}^+(i)=\sum_{j\in \mathcal{P}} H(x_{ij})$ with $x_{ij}=\predsingle_j-\predsingle_i$. Similarly, $\operatorname{rank}(i)$ is the rank of pixel $i$ among all predictions: $\operatorname{rank}(i)=\sum_{j\in \mathcal{P} \cup \mathcal{N}} H(x_{ij})$. These rank definitions rely on a step function, $H(\cdot)$, with a $\delta$-approximation around the step, as suggested by Chen et al.:
\begin{equation}
\label{eq:ap_step}
\small
H(x)=\left\{
\begin{aligned}
&0\, , & x < -\delta \\
&\frac{x}{2\delta}+0.5\, , & -\delta \leq x \leq \delta \\
&1\, , & \delta < x.
\end{aligned}
\right.
\end{equation}
Plugging these definitions into Eq. \ref{eqn:ap_loss} yields:
\begin{equation}
\begin{aligned}
\mathcal{L}_{\textrm{Rank}} & =1-\frac{1}{|\mathcal{P}|} \sum_{i \in \mathcal{P}} \frac{\sum_{j \in \mathcal{P}} H\left(x_{i j}\right)}{\sum_{j \in \mathcal{P} \cup \mathcal{N}} H\left(x_{i j}\right)}, \\ 
    & =\frac{1}{|\mathcal{P}|} \sum_{i \in \mathcal{P}} \sum_{j \in \mathcal{N}} L^{\textrm{Rank}}_{ij}
,
\end{aligned}
\end{equation} %\SK{$L(i,j)$ vs $L_{ij}$}
where $L^{\textrm{Rank}}_{ij}$, coined as the primary term, is defined as
\begin{equation}
\label{eq:ap_act}
L^{\textrm{Rank}}_{ij} =\frac{H\left(x_{i j}\right)}{\sum_{k \in \mathcal{P} \cup \mathcal{N}} H\left(x_{i k}\right)}.
\end{equation} 
%
%and $y_{ij}$ is defined as:

%\begin{equation}
%\label{eq:ap_score_diff}
%\begin{aligned}
%& \forall i, j,  \quad x_{i j} = -\left(s_i-s_j\right) \\
%& \forall i, j, \quad y_{i j}=\mathbf{1}_{i \in \mathcal{P} , j \in \mathcal{N}},
%\end{aligned}
%\end{equation} where $s_i$ is the $i$th pixel in the predicted edge map after flattening. 

%The details about how this formula is derived are presented in the appendix. appendixe ekle formülleri.

\subsection{Sorting Positives with their Uncertainties} %Multi-label Case: RS-Loss}
\label{sect:sorting}

Edge detection datasets contain annotations from multiple annotators, leading to (aleatoric) uncertainty in ground truth labels. In \Method, we favor accurate prediction of positive edge pixels with high certainty over those with low certainty. To this end, we use the sorting objective introduced by RS Loss \cite{oksuz2021rank}, where object detection hypotheses were sorted by their localization qualities. 

%While AP-Loss aims to rank each positive pixel above all negative pixels, Rank and Sort Loss, RS-Loss \cite{oksuz2021rank} also sorts positive pixels among themselves based on Intersection-over-union (IoU) for object detection. At this point, we use uncertainty-aware labels created from multi-label instead of IoU and we are thus assigning more importance to edge pixels that have lower uncertainty (lower uncertainty means labeled more by annotators). How we create uncertainty-aware labels is explained in detail at the end of this section. 

Following RS Loss \cite{oksuz2021rank}, we define our sorting loss as: %$\mathcal{L}_{\mathrm{Sort}}$ as follows:
\begin{equation}
\label{eq:rs_formula}
\mathcal{L}_{\mathrm{Sort}}=\frac{1}{|\mathcal{P}|} \sum_{i \in \mathcal{P}}\left(\ell_{\mathrm{Sort}}(i)-\ell_{\mathrm{Sort}}^*(i)\right) ,
\end{equation} where $\ell_{\mathrm{Sort}}(i)$ and $\ell_{\mathrm{Sort}}^*(i)$ are a summation of current sorting error and target sorting error, respectively. We define the sorting error $\ell_{\mathrm{Sort}}(i)$ as the average uncertainty of pixels with higher confidence than pixel $i$ (i.e., $H(x_{ij}) = 1$):
\begin{equation}
\label{eq:rs_current_err}
\begin{aligned}
\ell_{\mathrm{Sort}}(i) &= \frac{1}{\operatorname{rank}^{+}(i)} \sum_{j \in \mathcal{P}} H\left(x_{i j}\right) (1-\Unc_j),
\end{aligned}
\end{equation} 
where $\Unc_j \in [0,1]$ is the label certainty of pixel $j$ (explained in Section \ref{sect:uncertainty}), and $\operatorname{rank}^{+}(i)$ and $\mathrm{H}\left(x_{i j}\right)$ are as defined in Section \ref{sect:ranking}.
 
The target error $\ell_{\mathrm{Sort}}^*(i)$ is defined as the average uncertainty ($1-c_j$) over each positive pixel $j\in \mathcal{P}$ with higher confidence ($H(x_{ij}) = 1$) and higher certainty ($c_j \geq c_i$), calculated as:
\begin{equation}
\begin{aligned}
\ell_{\mathrm{Sort}}^*(i) &= \frac{\sum_{j \in \mathcal{P}} \mathrm{H}\left(x_{i j}\right)\left[\Unc_j \geq \Unc_i\right] (1-\Unc_j)}{\sum_{j \in \mathcal{P}} \mathrm{H}\left(x_{i j}\right)\left[\Unc_j \geq \Unc_i\right]} ,
\end{aligned}
\end{equation} where $[P]$, the Iverson Bracket, is 1 if $P$ is True and 0 otherwise.

We plug the definitions of $\ell_{\mathrm{Sort}}(i)$ and $\ell_{\mathrm{Sort}}^*(i)$ into $\mathcal{L}_{\mathrm{Sort}}$ to obtain the primary terms as in AP Loss as follows:
\begin{equation}
    L^{\textrm{Sort}}_{ij} = 
       \left(\ell_{\mathrm{Sort}}(i) - \ell_{\mathrm{Sort}}^*(i)\right) 
            \frac{H(x_{ij}) [\Unc_j > \Unc_i]}{ \sum_k H(x_{ik}) [\Unc_k > \Unc_i]}, 
\end{equation}
which is zero if $i,j\notin \mathcal{P}$. 

\subsection{Computing Pixel Uncertainties}
\label{sect:uncertainty}

Previous studies create labels for the multi-label datasets as follows: (i) Merge $n$ annotations $\{\mathbf{y}^{a}\}_{a=1}^n$ for one RGB image provided by $n$ annotators: ${\mathbf{y}^{*} \leftarrow \oplus_{a=1}^n  \mathbf{y}^{a}}$.
(ii) Binarize ${\mathbf{y}^{*}}$ using a chosen threshold $\tau$ to obtain the training target ${\mathbf{y}}$:
\begin{equation}
y_i = \begin{cases} 0 , \text{ for }  {y}^*_{i} < \tau  ,\\ 1, \text { otherwise , }\end{cases}
\end{equation} where $i \in \mathcal{P} \cup \mathcal{N}$.

This approach neglects both pixel-wise and label-wise uncertainties. Precise labeling of edges by hand is difficult. Most edge pixels annotated by humans do not overlap with low-level edges in RGB images. Hence, the evaluation procedure of edge detection tolerates localization error to match edges in predicted and ground-truth results \cite{dollar2014fast}. Therefore, a simple merging operation ignores this pixel-wise uncertainty in training.
Moreover, binarizing $\mathbf{y}^{*}$  not only leads to a loss of information about how many times edge pixels are labeled by  multiple annotators but also causes a loss of edges that are rarely labeled among these annotations.

To this end, we propose calculating a certainty map $\mathbf{c}$ which preserves the level of agreement among annotators (Algorithm \ref{alg:ual}). For this, first we take the pixel-wise logical-OR of the annotations: $\tilde{\mathbf{y}} = \operatorname{OR}_{a=1}^n {\mathbf{y}^a}$ ($\tilde{y}_i$ for pixel $i$). Then, we define the certainty $c_i$ for pixel $i$ as an average of how many annotations match an edge pixel in its $d$-vicinity in $\{\mathbf{y}^a\}_{a=1}^n$:
\begin{equation} \label{eq:uncertainty}
    \Unc_i = \frac{1}{n} \sum_{a=1}^n \textrm{CP}(\tilde{y}_i, \mathbf{y}^a, d),
\end{equation}
where $\textrm{CP}(\tilde{y}_i, \mathbf{y}^a, d)$ is a commonly used function in the edge detection benchmarks \cite{arbelaez2010contour,dollar2014fast} to find match predictions ($\tilde{y}_i\in \tilde{\mathbf{y}}$) with the ground truth ($\mathbf{y}^a$) within a $d$-Manhattan distance (CP: Correspond Pixels).

\begin{algorithm}

\hspace*{\algorithmicindent}  \textbf{Input: }
            - Set of $n$ annotations $\{\mathbf{y}^a\}_{a=1}^n$.\\
            \hspace*{1.7cm} - Maximum distance tolerance for overlap: ${d}$. \\
   \hspace*{\algorithmicindent}  \textbf{Output: }{Certainty map, $\mathbf{c}$ ($\Unc_i$: for pixel $i$)}.

    \begin{algorithmic}[1]
    %\State $n \leftarrow$  $\vert$$\mathrm{G}$$\vert$ \hfill \Comment{The number of different annotations.}
    \State $\tilde{\mathbf{y}} \leftarrow\mathbf{y}^{1} \operatorname{OR} \mathbf{y}^{2} \operatorname{OR} ... \operatorname{OR} \mathbf{y}^{n}$. \Comment{Combine annotations.}
    
    %\State  $\forall_{i} \forall_{j}$, $\mathrm{G}_{{UaL}_{ij}}$  $\leftarrow$ 0
    %\State  $\forall_{i} \forall_{j}$,  $temp_{ij}$ $\leftarrow$ 0
    \For{ each edge pixel $i$ in $\tilde{\mathbf{y}}$}
        \State $\Unc_i \leftarrow \frac{1}{n}\sum_{a=1}^n \textrm{CP}(\tilde{y}_i, \mathbf{y}^a, {d})$ \Comment{Eq. \ref{eq:uncertainty}.}
        %\For{$i \in \{1, ..., n\}$}
        %  \State num\_match $\leftarrow$ num\_match + CP($e$, $\mathbf{y}^{i}$, d)            
        %\EndFor
        %\State $u_e \leftarrow$ num\_match
    \EndFor
    %\State $\mathbf{u} \leftarrow \mathbf{u} / n$  \Comment{Normalization}

    %\State    $temp$ $\leftarrow$ $temp$ +  $G_i$
    % \For{j $\in$ $|\mathrm{G}|$}  
    %   \If{ i $\neq$ j} 
    %    \State $temp$ $\leftarrow$  $temp$ + CP($G_i$, $G_j$, $d$)
        
    %        \EndIf

    %    \EndFor
    %\State $temp$ $\leftarrow$ $temp$ $/$ $|G|$ 
    %\Comment{Normalization}
    %\For{j $\in$ $temp$.width}
    %    \Comment{Iteration over pixels}
    %    \For{k $\in$ $temp$.height}
    %        \If{  ${temp}_{jk} \neq 0 \textbf{ and } {\mathrm{G}_{UaL}}_{jk} =$ 0}
    %            \State $\mathrm{G}_{{UaL}_{jk}}$ $\leftarrow$  $temp$
    %        \EndIf
    %        \EndFor
    %        \EndFor
            
    %\State $temp$ $\leftarrow$ 0
    %\EndFor
    \caption{Computing the certainty map $\mathbf{c}$ of annotations.}
    \label{alg:ual}
    \end{algorithmic}
    
\end{algorithm}

%We do not give details of finding common pixels function $CP()$, because it is already used in standard evaluation procedure for edge detection.

\subsection{Overall Loss Function}

We combine the ranking and sorting components as follows: 
\begin{equation}\label{eq:overall_loss}
    \mathcal{L}_{\textrm{Overall}} = \mathcal{L}_{\textrm{Rank}} + \alpha\mathcal{L}_{\textrm{Sort}},
\end{equation}  
where $\alpha$ is a coefficient to control influence of sorting loss $\mathcal{L}_{\textrm{Sort}}$. In the case of a single-annotation dataset (i.e., $n=1$), we are unable to calculate an uncertainty map / the sorting loss $\mathcal{L}_\textrm{Sort}$; therefore, we set $\alpha$ to 0 for such datasets.

\subsection{Optimization}

The counting-based definitions of $\mathcal{L}_{\textrm{Rank}}$ and $\mathcal{L}_{\textrm{Sort}}$ prohibit an autograd-based calculation of $\partial \mathcal{L}_{o} / \partial p_i$ where $o \in \{\textrm{Rank, Sort}\}$. To address this issue, we follow the error-driven update mechanism of Chen et al. \cite{APloss-PAMI20} that approximates such a gradient as follows: 
\begin{equation}
\label{eq:ap_grad}
\begin{gathered}
\begin{aligned}
\frac{\partial \mathcal{L}_{o}}{ \partial p_i} = \sum_j L^{o}_{ji} t^o_{ji} - \sum_j L^{o}_{ij} t^o_{ij},
\end{aligned}
\end{gathered}
\end{equation}
where $t^o_{kl}=1$ if $y_k = 1$ and $y_l = 0$ for $o=\textrm{Rank}$;  $t^o_{kl}=1$ if $y_k = 1$ and $y_l = 1$ for $o=\textrm{Sort}$.
See Chen et al. \cite{APloss-PAMI20} for the derivation details.

%% file: sect_experiments.tex
\section{Experiments}

\subsection{Experimental Setup and Details}
\textbf{Datasets.} To evaluate our method, we carried out comprehensive experiments on three commonly used datasets: %NYUD-v2  \cite{silberman2012indoor}, BSDS500 \cite{arbelaez2010contour} and  Multi-cue \cite{mely2016systematic}.

\underline{BSDS500 \cite{arbelaez2010contour}} is one of the most popular datasets for evaluating edge detection. It officially contains 200 images for training, 100 images for validation, and 200 images for testing. Each image has 321 × 481 resolution and is manually labeled by at least five different annotators, which creates aleatoric uncertainty in the ground-truth labels. 

\underline{NYUDv2 \cite{silberman2012indoor}} contains 1449 paired RGB and Depth images with 576 × 448 resolution. %Although it is published with segmentation labels, it is widely used for edge detection. 
Edge detection labels are extracted using the boundaries of segmented objects. It has a single ground-truth edge map for each RGB image. It officially contains 381 images for training, 414 for validation, and 654 images for testing.

\underline{Multi-cue \cite{mely2016systematic}}  contains 100 natural images with 1280 × 720 resolution. Each image has six edge and five boundary labels. The dataset does not provide training and testing splits. We randomly allocated 80 images for training, and 20 images for testing following previous work \cite{pu2022edter,zhou2023treasure}.

\begin{table}[ht]
\centering
\resizebox{1\columnwidth}{!}{%
\begin{tabular}{|c|ccc|ccc|}
\hline
\multirow{2}{*}{Method}  & \multicolumn{3}{c|}{SS} & \multicolumn{3}{c|}{MS}  \\ \cline{2-7}
        &      ODS    & OIS   & AP    & ODS    & OIS   & AP    \\\hline\hline
Canny  \cite{canny1986computational} \kaynak{(PAMI’86)} & .611 & .676 & .520 & - & - & - \\
gPb-UCM \cite{arbelaez2010contour} \kaynak{(PAMI’10)} & .729 & .755 & .745 & - & - & -  \\
SCG \cite{xiaofeng2012discriminatively} \kaynak{(NeurIPS’12)} & .739 & .758 & .773 & - & - & -  \\
SE \cite{dollar2014fast} \kaynak{(PAMI’14)} & .743 & .764 & .800 & - & - & -  \\
OEF \cite{hallman2015oriented} \kaynak{(CVPR’15)} & .746 & .770 & .815 & - & - & -  \\\hline
DeepEdge \cite{bertasius2015deepedge} \kaynak{(CVPR’15)} & .753 & .772 & .807 & - & - & -  \\
DeepContour \cite{shen2015deepcontour} \kaynak{(CVPR’15)} & .757 & .776 & .790 & - & - & -  \\
HED \cite{xie2015holistically} \kaynak{(ICCV’15)} & .788 & .808 & .840 & - & - & -  \\
DeepBoundary \cite{kokkinos2015pushing} \kaynak{(ICLR’15)} & .789 & .811 & .789 & .803 & .820 & .848  \\
CEDN \cite{yang2016object} \kaynak{(CVPR’16)} & .788 & .804 & - & - & - & -  \\
RDS \cite{liu2016learning} \kaynak{(CVPR’16)} & .792 & .810 & .818 & - & - & -  \\
COB \cite{maninis2016convolutional} \kaynak{(ECCV’16)} & .793 & .820 & .859 & - & - & -  \\
AMH-Net \cite{xu2017learning} \kaynak{(NeurIPS’17)} & .798 & .829 & .869 & - & - & -  \\
RCF \cite{liu2017richer} \kaynak{(CVPR’17)} & .798 & .815 & - & - & - & -  \\
CED \cite{wang2017deep} \kaynak{(CVPR’17)} & .803 & .820 & .871 & - & - & -  \\
LPCB \cite{deng2018learning} \kaynak{(ECCV’18)} & .800 & .816 & - & - & - & -  \\
BDCN \cite{he2019bi} \kaynak{(CVPR’19)} & .806 & .826 & .847 & - & - & -  \\
DSCD \cite{deng2020deep} \kaynak{(ACMMM’20)} & .802 & .817 & - & - & - & -  \\
%LDC \cite{deng2021learning} (\kaynak{ACMMM’21)} & .799 & .816 & .837 & - & - & -  \\
FCL-Net \cite{xuan2022fcl} \kaynak{(NN’22)} & .807 & .822 & - & .816 & .833 & - \\
EDTER \cite{pu2022edter} \kaynak{(CVPR’22)} & \underline{.824} & \underline{.841} & .880 & \textbf{.840} & \textbf{.858} & .896  \\
UAED \cite{zhou2023treasure} \kaynak{(CVPR'23)} & \textbf{.829} & \textbf{.847} & \underline{.892} & \underline{.837} & \underline{.855} & .897 
\\
ACTD \cite{deng2023learning} \kaynak{(Neurocomp.'23)} & .817 & .836 & .839 & - & - & -  \\
\hline
\Method~(Ranking Only)  & .822 & .838 & .886 & .829 & .850 & \underline{.900} \\
\Method~(Ranking \& Sorting)  & \underline{.824} & .840 & \textbf{.895} & \underline{.837} & \underline{.855} & \textbf{.911} \\
\hline
\end{tabular}%
}
\caption{Quantitative results on BSDS dataset \cite{arbelaez2010contour}. SS and MS represent single-scale and multi-scale, respectively. %While Ours\textsubscript{R} represents using only ranking loss, Ours\textsubscript{R+S} represents using sorting and ranking losses together. 
Bold: Best result. Underline: Second-best results. %The best and second-best results are shown with bold and underlined texts, respectively. }
\label{tab:res_bsds}}
\end{table}

\begin{table}[ht]
\centering
\resizebox{1\columnwidth}{!}{%
\begin{tabular}{|c|ccc|ccc|}
\hline
\multirow{2}{*}{Method}  & \multicolumn{3}{c|}{SS+VOC} & \multicolumn{3}{c|}{MS+VOC} \\ \cline{2-7}
        &      ODS    & OIS   & AP    & ODS    & OIS   & AP   \\\hline\hline
LDC \cite{deng2021learning} (\kaynak{ACMMM’21)} & .812 & .826 & .857 & .819 & .834 & .860 \\
PiDiNet \cite{su2021pixel} \kaynak{(ICCV’21)}  & .807 & .823 & - & - & - & - \\
FCL-Net \cite{xuan2022fcl} \kaynak{(NN’22)}   & .815 & .834 & - & .826 & .845 & - \\
EDTER \cite{pu2022edter} \kaynak{(CVPR’22)}  & .832 & .847 & .886 & \textbf{.848} & \textbf{.865} & .903 \\
UAED \cite{zhou2023treasure} \kaynak{(CVPR'23)} & \textbf{.838} & \textbf{.855} & \textbf{.902} & \underline{.844} & \underline{.864} & \underline{.905}
\\
CHRNet \cite{elharrouss2023refined} \kaynak{(Pat. Rec.'23)} & .787 & .788 & .801 & .830 & .853 & .870 \\
ACTD \cite{deng2023learning} \kaynak{(Neurocomp.'23)}  & .821 & .837 & .850 & .826 & .842 & .854 \\
\hline
\Method~(Ranking Only)  & \underline{.833} & \underline{.848} & \underline{.901} & \underline{.844} & .860 & \textbf{.916} \\
\hline
\end{tabular}%
}
\caption{Quantitative results on BSDS dataset \cite{arbelaez2010contour} using extra Pascal Context Data \cite{everingham2010pascal} in training. SS and MS represent single-scale and multi-scale, respectively. The best and second-best results are shown with bold and underlined texts, respectively. %While Ours\textsubscript{R} represents using only ranking loss, Ours\textsubscript{R+S} is not applicable due to a lack of uncertainty in VOC dataset.  The best and second-best results are shown with bold and underlined texts, respectively.}
}
\label{tab:res_bsds_voc}
\end{table}

\noindent\textbf{Implementation and Training Details.} We use the MMSegmentation toolbox \cite{mmseg2020} to implement our method \Method. The official implementations of AP Loss \cite{chen2019towards} and RS Loss \cite{oksuz2021rank} are based on for-iterations, which lead to long training times. Therefore, we developed fully-vectorized implementations, boosting training $\sim$4.5x compared to iteration-based ones. While the fully-vectorized implementation requires huge GPU memory -- at least 45 GB for 320×320 input resolution with the base model of Swin-Transformer \cite{liu2021swin}, we also provide a semi-vectorized implementation which speeds up about 2.5× and requires 20 GB of memory for the same input and model settings.

For the NYUD-v2 dataset, we use the official training and validation sets for training. %like in \cite{xie2015holistically, pu2022edter}
We apply the following data augmentations like in previous studies \cite{xie2015holistically, pu2022edter}: (i) Scaling with 0.5 and 1.5, (ii) horizontal flip, and (iii) rotation with degrees 90, 180, and 270.

For the BSDS500 dataset, we use the official training and validation sets for training. By following the literature \cite{he2019bi,pu2022edter}, we apply horizontal flip, rotation with 16 different angles between [0,360], and scaling with 0.5 and 1.5. For experiments on using additional training data, we use the Pascal Context Dataset \cite{everingham2010pascal}. %For this setup, we simply use this dataset as the additional training set. 

For the Multi-cue dataset, we randomly choose 80 images for training, and the remaining images for testing. We repeat this procedure  3 times and report their average with standard deviation. We apply horizontal flip and rotation with 90, 180, and 360 degrees as the data augmentation.

For all datasets, we randomly crop RGB images to 320 $\times$ 320 resolution. We do not use any threshold to select positive pixels (i.e., $\tau = 0$). We take the $\delta$ parameter in $H(\cdot)$ (Eq. \ref{eq:ap_step}) as 0.4 for NYUD-v2 and 0.1 for BSDS and Multi-cue datasets in $\mathcal{L}_\textrm{Rank}$. In $\mathcal{L}_\textrm{Sort}$, $\delta$ is used as 0.1 for BSDS and Multi-cue datasets. The learning rate is 1e-6 for BSDS and 1e-5 for NYUD-v2 \& Multicue datasets. 

The number of iterations is 200k for NYUD-v2 and BSDS datasets and 140k for Multicue dataset because of the risk of overfitting.
Batch size is 1. The remaining settings of the optimizer are the same as the base model of Swin-Transformer \cite{liu2021swin} for the segmentation task. Also, we use pre-trained weights of ImageNet-22k \cite{deng2009imagenet} on 384x384 images. The sorting loss coefficient $\alpha$ (Eq. \ref{eq:overall_loss}) is used as 1 and 2 for Multicue and BSDS datasets, respectively.

For testing, we followed the standard edge detection evaluation protocol \cite{DollarICCV13edges} which includes non-maximum suppression (NMS) and skeleton-based edge thinning, respectively. Also, we set the maximum allowed distance ($d$ in Eq. \ref{eq:uncertainty}) as 0.011 for NYUD-v2, 0.0075 for BSDS500 \& Multicue datasets similar to previous studies \cite{xie2015holistically, he2019bi,pu2022edter}.

\noindent\textbf{Performance Measures.} We use the following three commonly used measures: Optimal Dataset Scale (ODS), Optimal Image Scale (OIS), and Average Prevision (AP). To calculate ODS, one threshold is used for binarizing the predicted edge maps in the dataset, whereas different thresholds are used for each image in OIS. Also, ODS and OIS are essentially $F_1$ scores, showing the best result at only one point on the precision-recall curve. On the other hand, AP gives an insight into the general performance of the model.

Unlike previous studies, we present uncertainty-aware results (UaR) using these measures for multi-label datasets. In this type of result, we separately report these three measures for different values of certainty (e.g., AP @ $c_i \geq 0.2$, which means AP if $c_i$ is thresholded at 0.2). %binarized certainty map $\mathbf{c}$ for test data:
%
%\begin{equation}
%c_i = \begin{cases} 0 , \text{ for }  c_i < \eta  ,\\ 1, \text { otherwise , }\end{cases}
%\end{equation} where $i \in \mathcal{P} \cup \mathcal{N}$, $\eta$ is certainty threshold and  $\in [0.0,1.0]$ . We choose n different $\eta$ for n annotations, i.e. $\eta$ $\in$ \{0.0, 1/n, 2/n, ..., 1.0\}.

\subsection{Experiment 1: Comparison with SOTA}
This section compares \Method~with the state-of-the-art (SOTA) methods including hand-crafted and deep-learning-based methods. We report the results of SOTA methods from their publications.\newline

\noindent \textbf{On BSDS500.}
We compare the performance of our model with SOTA methods, as shown in Tables \ref{tab:res_bsds} and \ref{tab:res_bsds_voc}, for different input and training data settings: Single Scale (SS), Multi-scale (MS), with additional Pascal-VOC dataset in Single Scale (SS+VOC), and with additional Pascal-VOC dataset in Multi-Scale (MS+VOC).
The results suggest that our method provides the best AP for all input and training data settings -- with only one exception at SS+VOC, where \Method~is 0.1 AP points behind UAED \cite{zhou2023treasure}. \Method's ODS and OIS performances are on par with others. Since the Pascal VOC dataset is not a multi-label dataset (i.e. no uncertainty), we cannot apply the sorting loss on it.
\newline

\noindent \textbf{On NYUD-v2.} We provide results for RGB images in Table \ref{tab:NYU_res}. Our \Method~achieves the best results in all three measures: 78.0\% ODS, 79.3\% OIS, 82.6\% AP. HHA and RGB-HHA results are reported in the supplementary material. \newline
%This increase in AP metric is what we expected because our loss functions aim to optimize overall performance (AP metric) rather than optimizing scores at one point (ODS and OIS metrics). 

%In addition, we have the best AP results  
%for HHA  and RGB-HHA inputs. We also have the %top results of ODS and OIS for RGB-HHA inputs.
\begin{table}[ht]
\centering
\resizebox{0.8\columnwidth}{!}{%
\begin{tabular}{|c|ccc|}
\hline
{Method}                         &  ODS   & OIS  & AP    \\ \hline\hline
\multicolumn{1}{|l|}{gPb-ucm \cite{arbelaez2010contour} \kaynak{(PAMI’11)}} & {.632} & {.661} & .562  \\ 
\multicolumn{1}{|l|}{Silberman et al. \cite{silberman2012indoor}\kaynak{\ (ECCV’12)}} & {.658} & {.661} & - \\ 
\multicolumn{1}{|l|}{gPb+NG \cite{gupta2013perceptual}\kaynak{ (CVPR’13)}} & {.687} & {.716} & .629  \\ 
\multicolumn{1}{|l|}{OEF \cite{hallman2015oriented}\kaynak{ (CVPR’15)}} & {.651} & {.667} & -  \\ \hline
\multicolumn{1}{|l|}{HED \cite{xie2015holistically}\kaynak{ (ICCV’15)}} & {.720} & {.734} & .734  \\ 
\multicolumn{1}{|l|}{RCF \cite{liu2017richer}\kaynak{ (CVPR’17)}} & {.729} & {.742} & -  \\ 
\multicolumn{1}{|l|}{AMH-Net \cite{xu2017learning}\kaynak{ (NeurIPS’17)}} & {.744} & {.758} & .765  \\ 
\multicolumn{1}{|l|}{LPCB \cite{deng2018learning}\kaynak{ (ECCV’18)}} & {.739} & {.754} & -  \\ 
\multicolumn{1}{|l|}{BDCN \cite{he2019bi}\kaynak{ (CVPR’19)}} & {.748} & {.763} & .770 \\
\multicolumn{1}{|l|}{PiDiNet \cite{su2021pixel}\kaynak{ (ICCV’21)}} & {.733} & {.747} & .715  \\ 
\multicolumn{1}{|l|}{EDTER \cite{pu2022edter}\kaynak{ (CVPR'22)}} & \underline{.774} & \underline{.789} & \underline{.797} \\ 
\multicolumn{1}{|l|}{ACTD \cite{deng2023learning}\kaynak{ (Neurocomp.'23)}} & {.762} & {.774} & -  \\ \hline
\multicolumn{1}{|l|}{\Method~(R)} & {\textbf{.780}} & {\textbf{.793}} & \textbf{.826} \\ \hline
\end{tabular}%
}
\caption{Quantitative comparisons on NYUD-v2 \cite{silberman2012indoor} for RGB images. %Ours\textsubscript{R} represents using only ranking loss. 
The best and second-best results are shown with bold and underlined texts, respectively. R: Ranking only.}
\label{tab:NYU_res}
\end{table}

\noindent \textbf{On Multicue.}
We compare our method with SOTA deep learning-based methods using both edge and boundary labels of Multi-cue dataset, as shown in Table \ref{tab:mult_res}. \Method~clearly outperforms all SOTA methods for edge and boundary detection in all metrics. On edge detection, it yields  +5.5, +4.3, and +2.3 percentage point improvements over the second-best results for ODS, OIS, and AP metrics, respectively. Similarly, on boundary, it gives +9.9, +9.5, and +6.8 percentage point improvements over the second-best results in ODS, OIS, and AP metrics, respectively.

\begin{table}[ht]
\centering
\resizebox{\columnwidth}{!}{%
\begin{tabular}{|c|c|c|c|}
\hline
Method    & ODS         & OIS         & AP          \\\hline\hline
Human   \cite{mely2016systematic} \kaynak{ (VR’16)}     & .750±.024  & -           & -           \\
Multicue \cite{mely2016systematic} \kaynak{ (VR’16)}     & .830±.002  & -           & -           \\
HED  \cite{xie2015holistically} \kaynak{  (ICCV’15)}   & .851±.014 & .864±.011 & -           \\
RCF   \cite{liu2017richer} \kaynak{  (CVPR’17)}   & .857±.004 & .862±.004 & -           \\
BDCN \cite{he2019bi}  \kaynak{  (CVPR’19)}   & .891±.001 & .898±.002 & .935±.002 \\
DSCD \cite{deng2020deep}  \kaynak{  (ACMMM’20)}  & .871±.007 & .876±.002 & -           \\
LDC \cite{deng2021learning}   \kaynak{  (ACMMM’21)}  & .881±.012 & .893±.011 & -           \\
PiDiNet \cite{su2021pixel} \kaynak{ (ICCV’21)}   & .855±.007 & .860±.005  & -           \\
FCL-Net \cite{xuan2022fcl}  \kaynak{ (NN’22)}     & .875±.005 & .880±.005  & -           \\
EDTER  \cite{pu2022edter} \kaynak{ (CVPR’22)}   & .894±.005 & .900±.003   & .944±.002 \\
UAED   \cite{zhou2023treasure} \kaynak{ (CVPR’23)}   & .895±.002 & .902±.001 & .949±.002 \\
CHRNet       \cite{elharrouss2023refined}    \kaynak{           (Pat. Rec.'23)} & .907\phantom{±.001}  & .922\phantom{±.001} & - \\ 
ACTD       \cite{deng2023learning}        \kaynak{       (Neurocomp.'23)}   & .890±.011  & .905±.009 & - \\ 
                           \hline
						   \Method~(${R}$)             &  \underline{.951±.002}	

           &       \underline{.953±.001}	    & \underline{.962±.004}         \\
\Method~(${R+S}$)            &    \textbf{.962±.003}
         & \textbf{.965±.003}
            &      \textbf{.973±.006}
      \\\hline
\multicolumn{1}{c}{ }& 	\multicolumn{1}{c}{\multirow{2}{*}{{(a) Edge}}} & \multicolumn{1}{l}{\multirow{1}{*}{}} \\
\multicolumn{1}{c}{ }& 	\multicolumn{1}{c}{\multirow{2}{*}{{}}} & \multicolumn{1}{l}{\multirow{1}{*}{}} \\
\hline
Method    & ODS         & OIS         & AP          \\\hline\hline   
Human  \cite{mely2016systematic} \kaynak{ (VR’16)}     & .760±.017  & -           & -           \\
Multicue \cite{mely2016systematic} \kaynak{ (VR’16)}     & .720±.014  & -           & -           \\
HED   \cite{xie2015holistically} \kaynak{  (ICCV’15)}   & .814±.011 & .822±.008 & .869±.015 \\
RCF  \cite{liu2017richer}  \kaynak{  (CVPR’17)}   & .817±.004 & .825±.005 & -           \\
BDCN  \cite{he2019bi}  \kaynak{ (CVPR’19)}   & .836±.001 & .846±.003 & .893±.001 \\
DSCD   \cite{deng2020deep} \kaynak{ (ACMMM’20)}  & .828±.003 & .835±.004 & -           \\
LDC  \cite{deng2021learning}  \kaynak{  (ACMMM’21)}  & .839±.012 & .853±.006 & -           \\
PiDiNet \cite{su2021pixel} \kaynak{ (ICCV’21)}   & .818±.003 & .830±.005  & -           \\
FCL-Net  \cite{xuan2022fcl} \kaynak{ (NN’22)}     & .834±.016 & .840±.016  & -           \\
EDTER  \cite{pu2022edter}  \kaynak{ (CVPR’22)}   & .861±.003 & .870±.004  & .919±.003 \\
UAED  \cite{zhou2023treasure}  \kaynak{ (CVPR’23)}   & .864±.004 & .872±.006 & .927±.006 \\
CHRNet       \cite{elharrouss2023refined}      \kaynak{         (Pat. Rec.'23)} & .859\phantom{±.001}  & .863\phantom{±.001} & - \\ 
ACTD       \cite{deng2023learning}        \kaynak{       (Neurocomp.'23)}   & .852 ±.004  & .863±.008 & - \\ \hline                  
\Method~(${R}$)          &  \underline{.954 ±.004}          &  \underline{.958± .005}           & \underline{.992 ±.002} \\
\Method~(${R+S}$)   &  \textbf{.963 ±.002}	
           & \textbf{.967±.002}	       & \textbf{.995±.001} \\\hline
\multicolumn{1}{c}{}& 	\multicolumn{1}{c}{\multirow{2}{*}{{\centering (b) Boundary}}} & \multicolumn{1}{l}{\multirow{1}{*}{}} \\
\multicolumn{1}{c}{}& 	\multicolumn{1}{c}{\multirow{2}{*}{{}}} & \multicolumn{1}{l}{\multirow{1}{*}{}} \\
\end{tabular}%
}
\caption{Quantitative results on Multicue Dataset \cite{mely2016systematic}. The best and second-best results are shown with bold and underlined texts, respectively. $R$: Ranking only. $R+S$: Ranking \& Sorting.}
\label{tab:mult_res}
\end{table}

\begin{table*}[ht]
\centering\setlength{\tabcolsep}{4pt}\scriptsize
%\resizebox{2.2\columnwidth}{!}{%
\begin{tabular}{|c||ccc|lll|lll|lll|lll|lll|}
\hline
    & \multicolumn{18}{c|}{Low Certainty (\textit{High Uncertainty}) $\xleftrightarrow{\makebox[5cm]{}}$ High Certainty (\textit{Low Uncertainty})}\\ \hline
   &  \multicolumn{3}{c|}{$\mathbf{c} > 0.0 $}                       & \multicolumn{3}{||c|}{$\mathbf{c} \geq 0.2 $}                                         & \multicolumn{3}{||c|}{$\mathbf{c} \geq 0.4 $}                                      & \multicolumn{3}{||c|}{$\mathbf{c} \geq 0.6 $}                                        & \multicolumn{3}{||c|}{$\mathbf{c} \geq 0.8 $}                                         & \multicolumn{3}{||c|}{$\mathbf{c} = 1.0 $}                                        \\ \hline
Method                                                       & \multicolumn{1}{c}{ODS}   & \multicolumn{1}{c}{OIS}   & AP    & \multicolumn{1}{||c}{ODS}   & \multicolumn{1}{c}{OIS}   & \multicolumn{1}{c|}{AP} & \multicolumn{1}{||c}{ODS}   & \multicolumn{1}{c}{OIS}   & \multicolumn{1}{c|}{AP} & \multicolumn{1}{||c}{ODS}   & \multicolumn{1}{c}{OIS}   & \multicolumn{1}{c|}{AP} & \multicolumn{1}{||c}{ODS}   & \multicolumn{1}{c}{OIS}   & \multicolumn{1}{c|}{AP} & \multicolumn{1}{||c}{ODS}   & \multicolumn{1}{c}{OIS}   & \multicolumn{1}{c|}{AP} \\ \hline \hline
HED   \cite{xie2015holistically} \kaynak{(ICCV’15)}               & \multicolumn{1}{c}{.788} & \multicolumn{1}{c}{.804} & .840 & \multicolumn{1}{||l}{.794}      & \multicolumn{1}{l}{.810}      &   .846                      & \multicolumn{1}{||l}{.812}      & \multicolumn{1}{l}{.829}      &    .864                     & \multicolumn{1}{||l}{.833}      & \multicolumn{1}{l}{.850}      &      .887                   & \multicolumn{1}{||l}{.852}      & \multicolumn{1}{l}{.868}      &     .902                    & \multicolumn{1}{||l}{.880}      & \multicolumn{1}{l}{.893}      &    .926                     \\ \hline
BDCN    \cite{he2019bi}         \kaynak{(CVPR’19)}              & \multicolumn{1}{c}{.815} & \multicolumn{1}{c}{.830} & .872 & \multicolumn{1}{||l}{.821}      & \multicolumn{1}{l}{.837}      &        .880                 & \multicolumn{1}{||l}{.842}      & \multicolumn{1}{l}{.859}      &       .899                  & \multicolumn{1}{||l}{.864}      & \multicolumn{1}{l}{.881}      &           .919              & \multicolumn{1}{||l}{.884}      & \multicolumn{1}{l}{.900}      &             .933            & \multicolumn{1}{||l}{.914}      & \multicolumn{1}{l}{.926}      &        .956                 \\ \hline
PiDiNet   \cite{su2021pixel}     \kaynak{(ICCV’21)}               & \multicolumn{1}{|c}{.807}     & \multicolumn{1}{c}{.823}     & .856    & \multicolumn{1}{||l}{.810}      & \multicolumn{1}{l}{.826}      &    .864                    & \multicolumn{1}{||l}{.829}      & \multicolumn{1}{l}{.846}      &     .883                   & \multicolumn{1}{||l}{.851}      & \multicolumn{1}{l}{.867}      &    .904                     & \multicolumn{1}{||l}{.871}      & \multicolumn{1}{l}{.887}      &     .923                    & \multicolumn{1}{||l}{.899}      & \multicolumn{1}{l}{.911}      &     .947                    \\ \hline
EDTER        \cite{pu2022edter}  \kaynak{(CVPR’22)}               & \multicolumn{1}{|c}{\textbf{.824}} & \multicolumn{1}{c}{\textbf{.841}} & .880  & \multicolumn{1}{||l}{.826}      & \multicolumn{1}{l}{.842}      &  .881                      & \multicolumn{1}{||l}{.848}      & \multicolumn{1}{l}{.864}      &    .899                     & \multicolumn{1}{||l}{.870}      & \multicolumn{1}{l}{.885}      &         .918                & \multicolumn{1}{||l}{.890}      & \multicolumn{1}{l}{.905}      &        .936             & \multicolumn{1}{||l}{.920}      & \multicolumn{1}{l}{\textbf{.932}}      &       .962                  \\ \hline
\hline
%CE                                                & -                     & \multicolumn{1}{|c}{.820} & \multicolumn{1}{c}{.837} & .871 & \multicolumn{1}{||l}{.827} & \multicolumn{1}{l}{.843} & .878                   & \multicolumn{1}{||l}{.848} & \multicolumn{1}{l}{.867} & .896                   & \multicolumn{1}{||l}{.871} & \multicolumn{1}{l}{.889} & .918                   & \multicolumn{1}{||l}{.890} & \multicolumn{1}{l}{.908} & .935                   & \multicolumn{1}{||l}{.920} & \multicolumn{1}{l}{.934} & .959                   \\ \hline
%CE + DICE                                         & -                     & \multicolumn{1}{|c}{.821} & \multicolumn{1}{c}{.836} & .872 & \multicolumn{1}{||l}{.827} & \multicolumn{1}{l}{.843} & .879                   & \multicolumn{1}{||l}{.848} & \multicolumn{1}{l}{.865} & .897                   & \multicolumn{1}{||l}{.871} & \multicolumn{1}{l}{.888} & .919                   & \multicolumn{1}{||l}{.891} & \multicolumn{1}{l}{.907} & .937                   & \multicolumn{1}{||l}{.918} & \multicolumn{1}{l}{.933} & .960                   \\ \hline
{\Method~}(${R}$)                                             & \multicolumn{1}{|c}{.822} & \multicolumn{1}{c}{.838} & .886 & \multicolumn{1}{||l}{.828} & \multicolumn{1}{l}{.844} & .890                   & \multicolumn{1}{||l}{.849} & \multicolumn{1}{l}{.866} & .909                   & \multicolumn{1}{||l}{.870} & \multicolumn{1}{l}{.886} & .924                   & \multicolumn{1}{||l}{.889} & \multicolumn{1}{l}{.904} & .939                   & \multicolumn{1}{||l}{.916} & \multicolumn{1}{l}{.923} & .962                   \\\cdashline{2-19}[.4pt/2pt]
{\Method~}(${R+S}$)                                             & \multicolumn{1}{|c}{\textbf{.824}} & \multicolumn{1}{c}{.840} & \textbf{.895} & \multicolumn{1}{||l}{\textbf{.831}} & \multicolumn{1}{l}{\textbf{.847}} & \textbf{.900}                   & \multicolumn{1}{||l}{\textbf{.854}} & \multicolumn{1}{l}{\textbf{.870}} & \textbf{.918}                   & \multicolumn{1}{||l}{\textbf{.877}} & \multicolumn{1}{l}{\textbf{.892}} & \textbf{.936}                   & \multicolumn{1}{||l}{\textbf{.897}} & \multicolumn{1}{l}{\textbf{.911}} & \textbf{.951}                   & \multicolumn{1}{||l}{\textbf{.926}} & \multicolumn{1}{l}{.929}      & \textbf{.970}                  \\ \hline
\end{tabular}%
%}

\caption{Uncertainty-aware results on BSDS dataset. All SOTA results are computed using official weights. Also, $c$ represents the certainty map mentioned in Algorithm \ref{alg:ual}. While case $c > 0.0$ contains all ground-truth edges in all labels, case $c = 1.0$ contains only ground-truth edges that are labeled in all labels. $R$: Only ranking, $R+S$: Ranking \& Sorting.}
\label{tab:bdcn_unc}
\end{table*}

\subsection{Experiment 2: Uncertainty-aware Evaluation}

This section provides uncertainty-aware evaluation (UaR) of \Method~and SOTA methods which published their weights on the BSDS dataset as shown in Table \ref{tab:bdcn_unc}. As most studies published their models for only the BSDS dataset,  we could not do a similar analysis on the Multi-cue dataset.

For the BSDS dataset, we define six uncertainty levels. For example, level $\mathbf{c}$ $>$ 0.0 contains all ground-truth edges (high + low uncertainties) in all labels. Therefore, this case is equivalent to the standard evaluation of edge detectors. On the other hand,  level $\mathbf{c} = 1.0$ (lowest uncertainty) contains only ground-truth edges which are labeled by all annotators.

The results in Table \ref{tab:bdcn_unc} suggest a similar behavior for all models: They are more successful on low-uncertainty edges than high-uncertainty ones. While the uncertainty level is increased (i.e., $\mathbf{c}$ goes from 1.0 to 0.0), the scores of ODS, OIS, and AP measures decrease.

Our method \Method~gives better results when the uncertainty level is decreased. For example, our method gives nearly the same performance with EDTER \cite{pu2022edter} for the ODS and OIS measures in level $\mathbf{c}$ $>$ 0.0, and the performance of our model increases, as the uncertainty level decreases. This can be explained by our sorting task because it assigns higher importance to lower uncertainty pixels.

This set of experiments and our uncertainty-aware results (UaR) reveal that edge detection models give a better performance when the uncertainty levels are decreased and most of the performance loss is caused by higher uncertainty pixels, especially labeled by only one annotator. Therefore, focusing on those pixels can provide more performance gain for the future study. UaR results of Multi-cue dataset are reported in the supplementary material.

\subsection{Experiment 3: Ablation Analysis}

\textbf{Different loss functions.} To show the effectiveness of our method, we perform step-by-step comparisons by including class-balanced cross-entropy and the combination of cross-entropy and dice losses \cite{deng2018learning} with the same model and input settings on NYUD-v2, BSDS, and Multi-cue datasets. The results in Table \ref{tab:abl_stdy} suggest that ranking systematically improves the results over the commonly used loss functions in the literature. Moreover, we see that, whenever feasible, using the sorting component provides gains compared to only using ranking. %our method outperforms these commonly used loss functions in all metrics for all datasets. Also, the proposed sorting task gives extra performance gain to the ranking task.

\begin{table}[ht]
\centering
\resizebox{\columnwidth}{!}{%
\begin{tabular}{|c|c|c|c|c|}
\hline
Dataset                                                             & Loss & ODS  & OIS  & AP   \\ \hline\hline
\multirow{3}{*}{NYUD-v2}                                                     & {CE}$_{CB}$           & .775         & .789         & .802         \\ \cdashline{2-5}[.4pt/2pt] 
                                                                             & CE + DICE     & .779         & .791         & .807         \\ \cdashline{2-5}[.4pt/2pt] 
                                                                             & {\Method~}(${R}$) & \textbf{.780}         & \textbf{.793}        & \textbf{.826}        \\ \hline
\hline
\multirow{4}{*}{BSDS}                                                        & {CE}$_{CB}$           & .820         & .831         & .871         \\ \cdashline{2-5}[.4pt/2pt]  
                                                                             & CE + DICE     & .821         & .836         & .872         \\ \cdashline{2-5}[.4pt/2pt]  
                                                                             & {\Method~}(${R}$) & .822         & .838         & .886         \\ \cdashline{2-5}[.4pt/2pt] 
                                                                             & {\Method~}(${R+S}$) & \textbf{.824}         & \textbf{.840}        & \textbf{.895}         \\ \hline\hline
\multirow{4}{*}{\begin{tabular}[c]{@{}c@{}}Multicue\\ Edge\end{tabular}}     & {CE}$_{CB}$           & .926±.006   & .926±.007   & .880±.001    \\ \cdashline{2-5}[.4pt/2pt] 
                                                                             & CE + DICE     & .930±.005   & .931±.004   & .883±.006   \\ \cdashline{2-5}[.4pt/2pt]  
                                                                             & {\Method~}(${R}$) & .951±.002               &   .953±.001             &  .962±.004              \\ \cdashline{2-5}[.4pt/2pt] 
                                                                             & {\Method~}(${R+S}$) &  \textbf{.962±.003}        & \textbf{.965±.003}              & \textbf{.973±.006}              \\ \hline\hline
\multirow{4}{*}{\begin{tabular}[c]{@{}c@{}}Multicue\\ Boundary\end{tabular}} & {CE}$_{CB}$           & .934±.004   & .943±.003   & .981±.003   \\ \cdashline{2-5}[.4pt/2pt]  
                                                                             & CE + DICE     & .947±.004   & .954±.004   & .984±.004   \\ \cdashline{2-5}[.4pt/2pt] 
                                                                             & {\Method~}(${R}$) & .954 ±.004     & .958± .005   & .992 ±.002  \\ \cdashline{2-5}[.4pt/2pt] 
                                                                             & {\Method~}(${R+S}$) & \textbf{.963 ±.002 } & \textbf{.967±.002} & \textbf{.995±.001} \\ \hline
\end{tabular}%
}
\caption{Quantitative comparison between our method and commonly used loss functions in edge detection. Bold numbers show the best results for corresponding columns and datasets. While Ours\textsubscript{R} represents using only ranking loss, $R$: Ranking. $R+S$: Ranking \& Sorting. CE\textsubscript{CB}: Class-balanced cross-entropy. }
\label{tab:abl_stdy}
\end{table}

\textbf{Our certainty map computation vs. label averaging.} We also conduct an experiment to show the efficiency of our certainty map $\mathbf{c}$ on BSDS dataset. In this experiment, as shown in Table \ref{tab:comp_label}, we use pixel-wise averaging and use it as the certainty map ($\mathbf{c} \leftarrow \sum_{a} \mathbf{y}^a$). In this experiment, all settings are the same except for the labels used. Our proposed certainty map $\mathbf{c}$ gives better performance than the standard labels obtained by pixel-wise averaging in all metrics because we consider pixel-wise uncertainty using a dataset-specific distance toleration like an evaluation protocol of edge detection. In this way, we overlap edges in multi-label that do not overlap in case of pixel-wise averaging and these overlaps provide labels that are more informative to the sorting task in terms of uncertainty.

\textbf{Prioritize Uncertain Pixels.} Our sorting loss prioritizes low-uncertainty pixels while Zhou et al. \cite{zhou2023treasure} give higher priority to high-uncertainty edges. To validate our approach, we integrate their uncertainty weighting scheme into our loss function as follows: $(1-c_i) \cdot \mathcal{L}(i)$ for pixel $i$. The results in Table \ref{tab:comp_label} suggest that this significantly drops the performance of our model. Therefore, giving more importance to low-uncertainty edges provides a better performance in \Method.

\begin{table}[ht]
\centering
\resizebox{1\columnwidth}{!}{%
\begin{tabular}{|c|c|c|c|}
\hline
Label Type                        & ODS            & OIS            & AP             \\ \hline\hline
Label Averaging as Certainty ($\mathbf{c} \leftarrow \sum_{a} \mathbf{y}^a$)    & 0.818          & 0.831          & 0.880          \\ \hline
Loss Weighting with Uncertainty ($1-c_i$)\hfill & 0.789          & 0.813          & 0.849          \\ \hline
\Method \hfill & \textbf{0.824} & \textbf{0.840} & \textbf{0.895} \\ \hline
\end{tabular}%
}
\caption{Comparison with pixel-wise averaging of labels as certainty, and uncertainty-weighted loss.}
\label{tab:comp_label}
\end{table}

\subsection{Experiment 4: Qualitative Comparison}
\begin{figure}[ht]
\centering
{
\scriptsize
\renewcommand{\arraystretch}{0.2}
\setlength{\tabcolsep}{3pt}
\newcommand{\SKALA}[0]{0.125}
\begin{center}
\begin{tabular}{ |m{1.3cm}|m{2cm} m{2cm} m{2cm} |}

\hline

\CStart{}Input\CEnd{}         & 
    \includegraphics[scale=\SKALA]{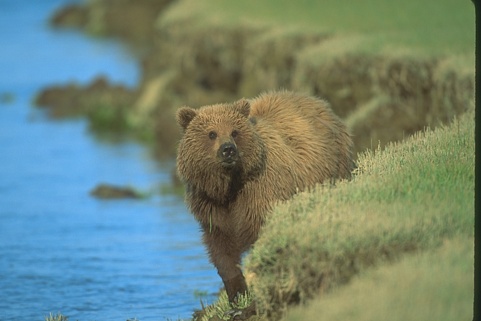} &
    \includegraphics[scale=\SKALA]{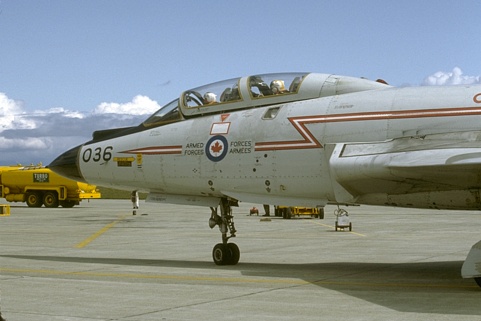} &
    \includegraphics[scale=\SKALA]{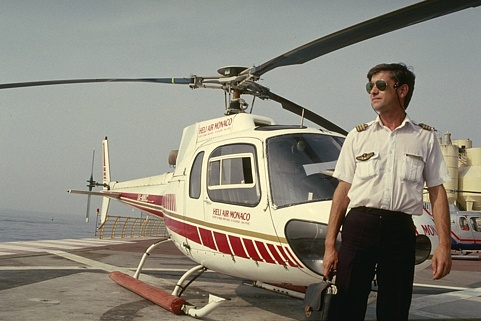} \\\hline

\CStart{}Ground-truth\CEnd{} &
    \includegraphics[scale=\SKALA]{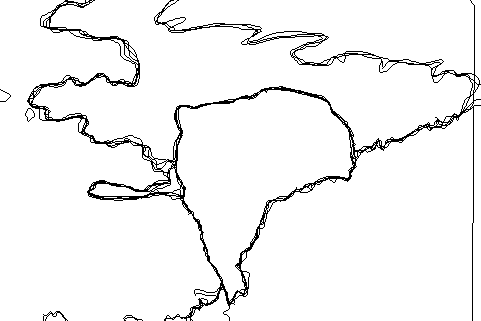} &
    \includegraphics[scale=\SKALA]{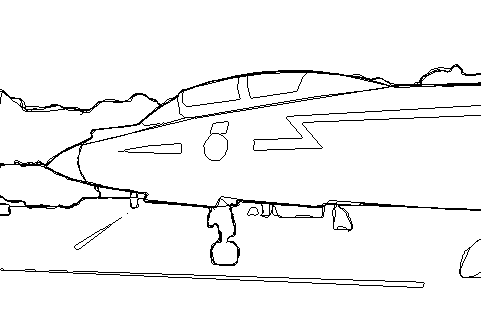} &
    \includegraphics[scale=\SKALA]{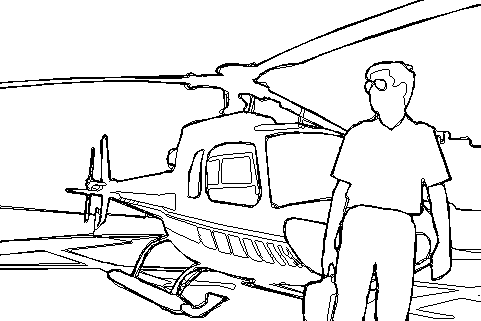} \\\hline

\CStart{}HED \cite{xie2015holistically}\CEnd{} &
    \includegraphics[scale=\SKALA]{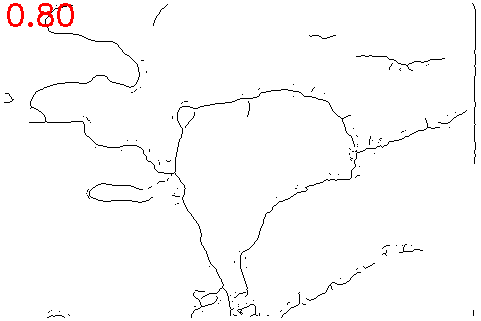} &
    \includegraphics[scale=\SKALA]{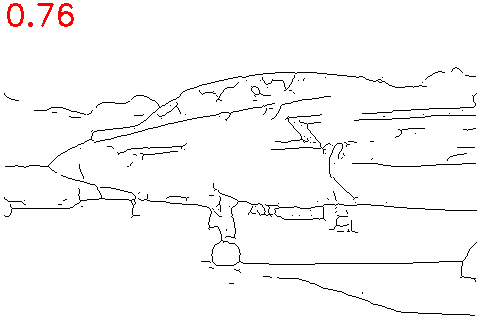} &
    \includegraphics[scale=\SKALA]{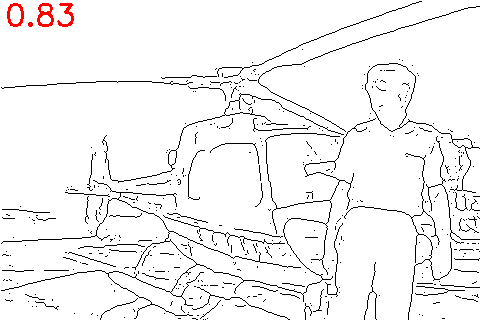} \\\hline

\CStart{}BDCN \cite{he2019bi}\CEnd{} &
    \includegraphics[scale=\SKALA]{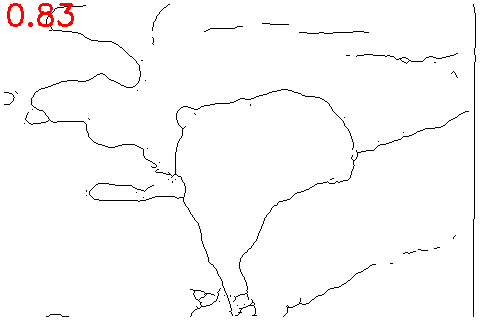} &
    \includegraphics[scale=\SKALA]{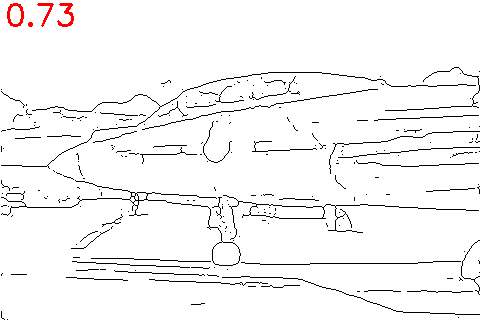} &
    \includegraphics[scale=\SKALA]{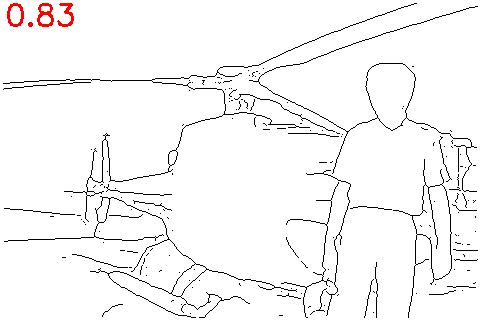} \\\hline

\CStart{}PiDiNet \cite{su2021pixel}\CEnd{} &
    \includegraphics[scale=\SKALA]{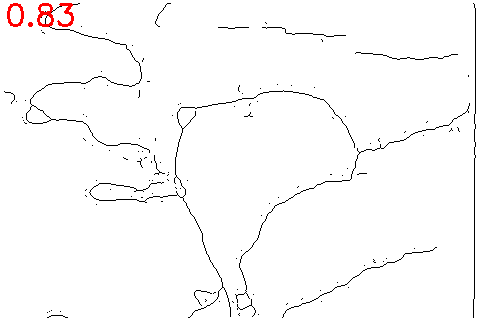} &
    \includegraphics[scale=\SKALA]{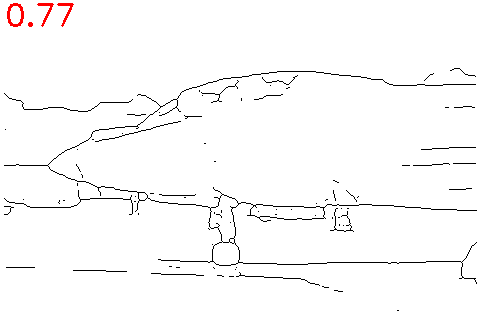} &
    \includegraphics[scale=\SKALA]{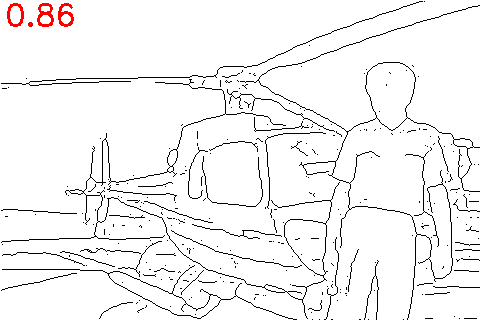} \\\hline

\CStart{}EDTER \cite{pu2022edter}\CEnd{} &
    \includegraphics[scale=\SKALA]{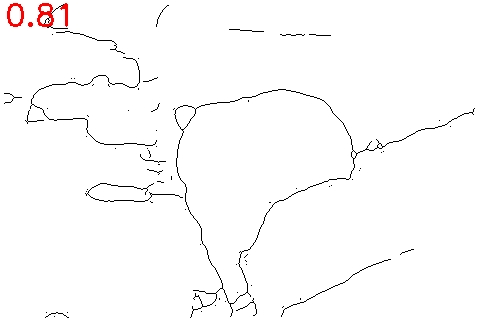} &
    \includegraphics[scale=\SKALA]{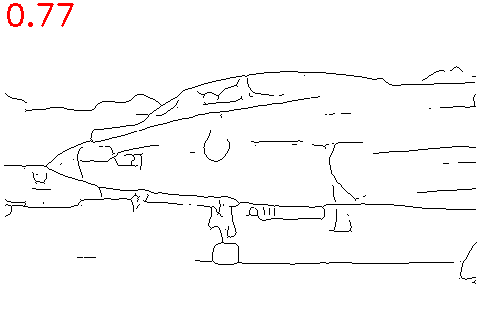} &
    \includegraphics[scale=\SKALA]{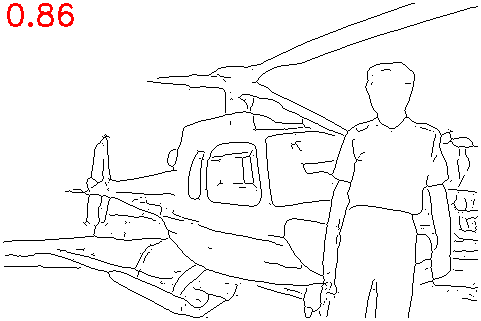} \\\hline
    
\CStart{}\Method~ \\ (Ours)\CEnd{}  &
    \includegraphics[scale=\SKALA]{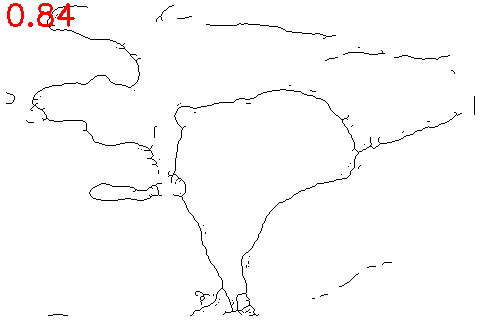}  &
    \includegraphics[scale=\SKALA]{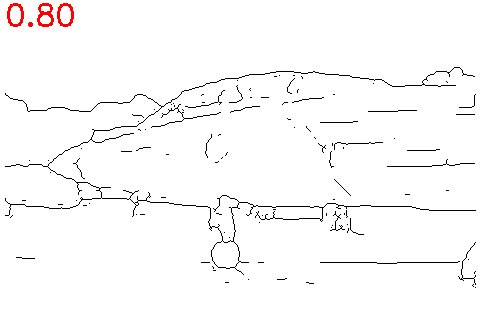}  &
    \includegraphics[scale=\SKALA]{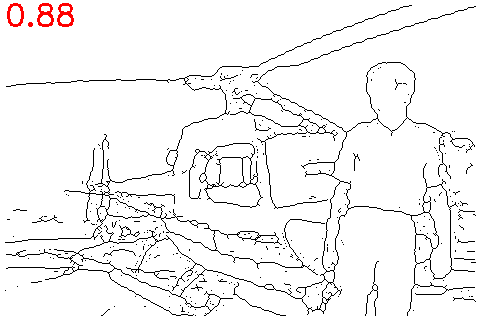}  \\\hline
\end{tabular}
\end{center}
}

 \caption{Qualitative results on BSDS dataset. All outputs are obtained after the post-processing step. Red: OIS scores. \label{fig:qualitative_results}}
%  \label{fig:vis_bsds}
\end{figure}

We also provide a qualitative comparison between SOTA models and ours on BSDS dataset, as shown in Figure \ref{fig:qualitative_results}. These results are presented after the post-processing step using thresholds of OIS measure. While our model detects prominent edges better, it may miss low-level details.  See the Supp. Mat. for more results.

\subsection{Experiment 5: Running Time Comparisions}
Although ranking-based loss functions give superior performance \cite{APloss-PAMI20, oksuz2021rank}, they suffer from long training time due to for-loops in their official implementations. Our vectorized implementations solve this problem %and boost training, 
as shown in Table \ref{tab:run_comp}. For a fair comparison, we test their execution time using loss functions with random tensors. Our vectorized and semi-vectorized implementations of $\mathcal{L}_{\textrm{Rank}}$ and $\mathcal{L}_{\textrm{Sort}}$ are almost 100 and 90 times, respectively, faster than their official implementations. However, when they are integrated into models, we observe less performance increase due to other bottlenecks such as forward/backward in models.

\begin{table}[ht]
\centering
\resizebox{1\columnwidth}{!}{%
\begin{tabular}{|l|l|r|}
\hline
Implementations                  & Loss      & Exec. Time (msec/img)   \\ \hline\hline
             -                    & CE        &   0.32             \\ \cdashline{2-3}[0.4pt/2pt]
             -                    & CE+DICE   &   0.39             \\ \hline
             \hline
\multirow{2}{*}{For-iterations}  & {\Method~}(${R}$)   &   142.79            \\ \cdashline{2-3}[0.4pt/2pt]
                                 & {\Method~}(${R}$ + ${S}$) & 1530.22               \\ \hline
\multirow{2}{*}{Semi-vectorized} & {\Method~}(${R}$)   &  2.90              \\ \cdashline{2-3}[0.4pt/2pt]
                                 & {\Method~}(${R}$ + ${S}$) & 16.73               \\ \hline
\multirow{2}{*}{Vectorized}      & {\Method~}(${R}$)   &  1.00             \\ \cdashline{2-3}[0.4pt/2pt]
                                 & {\Method~}(${R}$ + ${S}$) & 14.09              \\ \hline
\end{tabular}%
}
\caption{Execution time comparisons (avg. of 100 runs).}
\label{tab:run_comp}
\end{table}

%\begin{figure*}
%\centering
%\includegraphics[width=0.9\linewidth]{figures/vis_bsds.pdf}
%  \caption{Qualitative comparisons on BSDS dataset. Although our method produces edge maps with some background pixels, these are easily eliminated in the post-processing step as shown in the last two columns.}
%  \label{fig:vis_bsds}
%\end{figure*}

%% file: sect_conclusion.tex
\section{Conclusion}
We presented a ranking-based solution for object contour detection, or as widely referred to as edge detection, to address both the imbalance problem between positive (edge) and negative (non-edge) classes and uncertainty arising from disagreement between different annotators. Our solution contained two novel loss functions: One for ranking positive edge pixels over negatives, and another for sorting positive pixels with respect to their edge certainties. Our extensive experiments show the efficiency of our proposal  on NYUD-v2, BSDS, and Multicue datasets. \Method~can train an edge detector maximize AP (via its the ranking loss). Our paper paves the way for optimizing other objectives and making all post processing steps as part of the training process.

%\noindent \textbf{Future Work.} Although the proposed method outperforms SOTA models, it produces low-quality edge maps. We plan to improve our model in a way that does not require post-processing.

%\noindent \textbf{Limitations.} \Method~produces edge maps with some background pixels, although they can be easily eliminated after the post-processing step.